\documentclass[letterpaper, 10 pt, conference]{ieeeconf}
\IEEEoverridecommandlockouts
\usepackage{cite}

\usepackage{graphicx}
\usepackage{multirow}
\usepackage{dblfloatfix}

\usepackage[bookmarks=true]{hyperref}
\hypersetup{%
  breaklinks,
  bookmarksnumbered=true,
  bookmarksopen=true,
  colorlinks=true,
  linkcolor=black,
  urlcolor=black,
  citecolor=black
}

\usepackage{subcaption}
\usepackage{amsmath}
\usepackage{amsfonts}
\usepackage{comment}
\usepackage{algorithm}
\usepackage[noend]{algpseudocode}
\usepackage{bbding}

\usepackage{array}

\makeatletter
\newcommand{\thickhline}{%
    \noalign {\ifnum 0=`}\fi \hrule height 1.5pt
    \futurelet \reserved@a \@xhline
}
\newcolumntype{"}{@{\hskip\tabcolsep\vrule width 1.5pt\hskip\tabcolsep}}
\makeatother

\renewcommand{\arraystretch}{1.5}

\begin{document}
\bstctlcite{IEEEexample:BSTcontrol} %This line, along with the text at the top of the .bib file, shortens citations to use et al

\title{\LARGE \bf Duty Factor Predicts Robust Constrained\\ Quadrupedal Locomotion Across Gait Types}
\begin{comment}
    Possible titles:
    Designing Safe Quadrupedal Gaits
    The Effect of Gait Parameters on Safe Quadrupedal Robot Locomotion
\end{comment}

\author{James Zhu$^{1,2}$, David Ologan$^2$, George Ortiz$^2$, Thomas Chun Fai Lee$^3$, Selvin Garcia Gonzalez$^2$,\\ Ardalan Tajbakhsh$^2$,  Pinhas Ben-Tzvi$^1$, and Aaron M. Johnson$^{2,3}$%
    \thanks{This material is based upon work supported by the National Science Foundation
    under grant \#CMMI-1943900, the
GEM Consortium Fellowship, as well as the Assistant Secretary of the Army for Acquisition, Logistics, and Technology (ASA) ALT FUZE Contracting Office under Contract No.\ W51701-26-C-A154, in collaboration with FieldAI.}%
    \thanks{$^1$ Department of Electrical and Computer Engineering, University of Miami, Coral Gables, FL, USA, \texttt{jxz1636@miami.edu}}
    \thanks{$^2$ Department of Mechanical Engineering, $^3$ Robotics Institute, Carnegie Mellon University, Pittsburgh, PA, USA, \texttt{amj1@andrew.cmu.edu}}%
}

\begin{comment}
\author{Authors Anonymized for Review}
\end{comment}
% make the title area
\maketitle
\thispagestyle{empty}
\pagestyle{empty}

%%%%%%%%%%%%%%%%%%%%%%%%%%%%%%%%%%%%%%%%%%%%%%%%%%%%%%%%%%%%%%%%%%%%%%%%%%%%%%%%

\begin{abstract}

Quadrupedal robots are increasingly deployed in environments where locomotion must remain robust to disturbances and constrained terrain. 
Gait type, such as walking or trotting, is commonly used to characterize quadrupedal locomotion. 
However, gait type does not uniquely define locomotion, as parameters such as duty factor, speed, and stance width can vary within a single gait type. In this work, we investigate the relationship between these gait parameters using three distinct quadrupedal locomotion control approaches. 
First, using whole-body trajectory optimization with LQR feedback, we show that duty factor is a stronger predictor of local error convergence than nominal gait type. 
Second, we investigate duty-factor selection with a learned locomotion controller, suggesting how duty factor may serve as a low-dimensional parameter for adapting locomotion robustness in narrow-terrain environments. 
Finally, we show that these trends persist under a centroidal model-predictive control framework and validate them through narrow-terrain experiments on a physical quadruped. 
These results show that duty factor provides a simple and effective basis for understanding and selecting robust quadrupedal locomotion across gait types and control architectures.

\end{abstract}

%\begin{keywords}
%Gait Parameters, Robustness, Legged Robots
%\end{keywords}

%\vspace{-1em}
\section{Introduction}

The growing deployment of quadrupedal robots in real-world environments requires locomotion that remains robust to disturbances, uncertainty, and terrain constraints. Common deployments include construction sites \cite{halder_construction_2023}, offshore electrical substations \cite{gehring_offshore_2021}, and mine tunnels \cite{miller_darpa_2020}, where robots may encounter steep slopes, narrow pathways, and other challenging terrain. In these environments, robots must maintain reliable motion in the presence of disturbances and errors. A range of methods have therefore been developed to improve the robustness of legged locomotion \cite{zhu2023convergent,hammoud2021impedance,drnach_robust_2021,miki_robust_2022,belvedere2026contact}. 
However, these approaches often rely on non-trivial choices of locomotion and controller parameters. 
For example, trajectory optimization methods commonly require contact sequences, contact timings, gait periods, and speeds to be specified in advance \cite{hammoud2021impedance,manchester_ditrel_2019}.
Similarly, learning-based approaches also often fix foot contact sequences and timings to simplify policy training \cite{yu2026discovery,aractingi2023controlling}.
Understanding which gait parameters most directly influence locomotion performance is therefore important for both designing and adapting robust locomotion strategies.

Gait type is one of the most common abstractions used to characterize quadrupedal locomotion. Walking, trotting, pacing, and bounding are distinguished primarily by the relative timing of contacts between the legs. 
Many works have found correlations between gait type and performance, observing that different gaits are energetically more efficient at different speeds \cite{hoyt_horse_1981,Xi_gait_selection_2016, fu_emergent_gaits_2022,yu2026discovery}.
However, gait type does not uniquely determine locomotion behavior. 
A single gait type can be executed across a range of speeds, duty factors, and stance widths, and these parameters can substantially affect locomotion performance. Consequently, existing analyses may conflate the effects of gait type with the effects of the parameters used to realize each gait. This raises a fundamental question: \emph{which properties of a gait actually determine its robustness?}

\begin{figure}[t]
    \centering
    \includegraphics[width=0.95\linewidth]{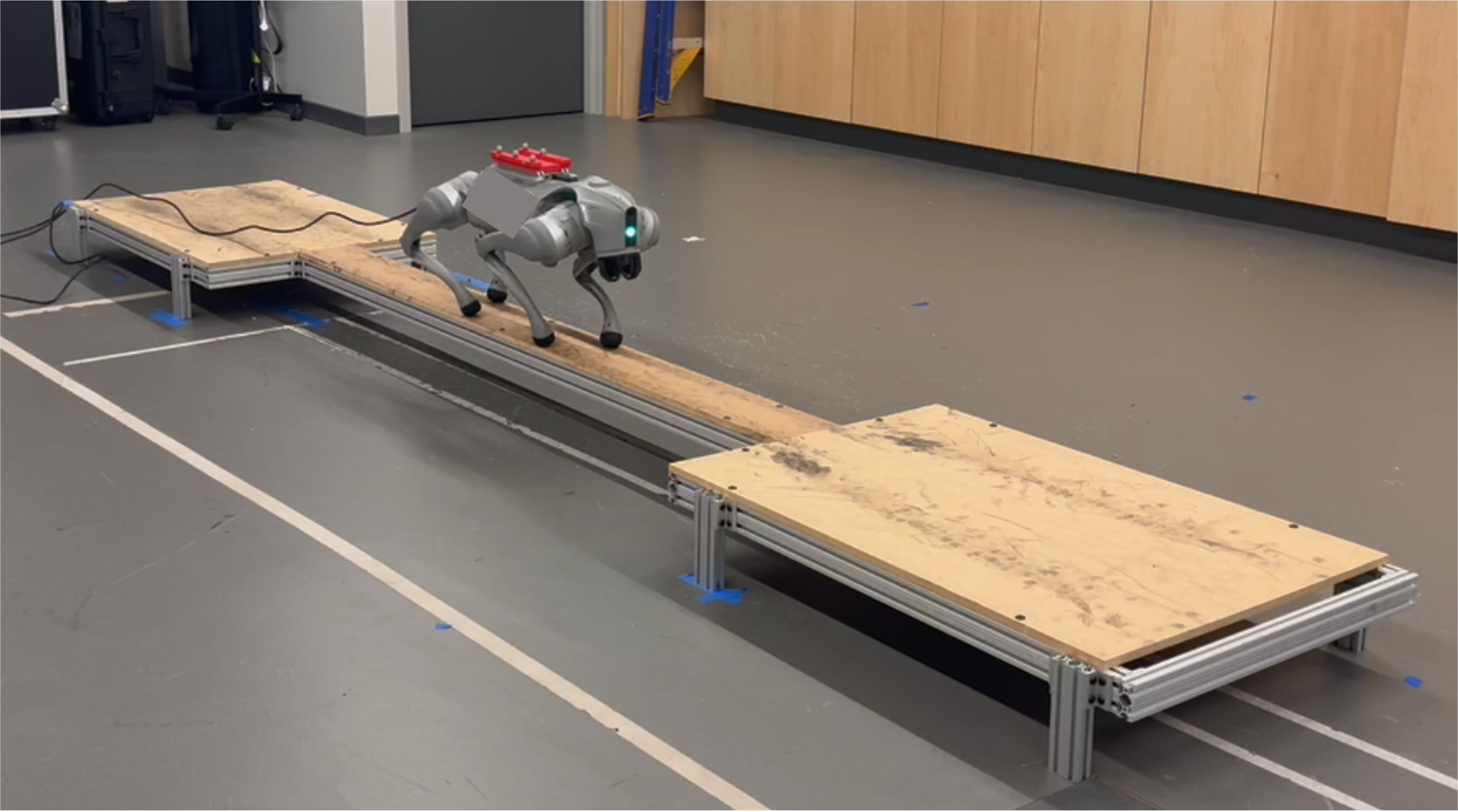}
    \caption{In this work, we demonstrate that the robustness of quadruped locomotion improves with an increase in duty factor. This finding is validated with the first demonstration of a quadruped robot walking across a narrow beam without the aid of external mechanisms like a reaction wheel.}
    \label{fig:beamwalking}
\end{figure}

Rather than treating nominal gait type as the sole determinant of locomotion performance, we investigate whether the underlying continuous gait parameters provide a more informative description of robustness. In particular, we focus on duty factor (DF), which specifies the fraction of a gait cycle that each foot remains in contact with the ground. Duty factor directly affects the distribution of ground reaction forces and can therefore influence how perturbations are transmitted through the robot and corrected by the controller. 
%Duty factor can also shared across different gait types, providing a natural basis for comparing walking and trotting while controlling for one of their key differences.

The key result of this paper is that duty factor is a stronger predictor of local error convergence than nominal gait type. 
Saltation-based sensitivity analysis \cite{kong2023saltation,belvedere2026contact} provides a theoretical framework to compute the robustness of a trajectory to disturbances, which we validate through a number of empirical trials in simulation and hardware.
First, using whole-body trajectory optimization with LQR feedback, we systematically vary gait type, duty factor, speed, and stance width and evaluate their effects on a local convergence metric. 
Using a linear regression model, we find that gait type is not significantly correlated with convergence when evaluated at matched duty factors. 
This result is particularly notable because the conventional distinction between walking and trotting would suggest substantially different locomotion performance \cite{humphreys2025learning,shafiee2024viability}. 
%Instead, our results indicate that duty factor captures a more fundamental dimension of convergence behavior observed in these gaits.

We further examine whether duty factor can be used as a low-dimensional variable for autonomous gait adaptation. 
We develop a learned locomotion controller that selects duty factor according to terrain width conditions. 
This experiment connects the relationship between duty factor and robustness to a practical mechanism for adapting legged locomotion.

Finally, using a separate centroidal model-predictive control framework with foothold planning, we evaluate walking and trotting gaits across different duty factors and speeds with a real quadruped robot navigating across a narrow beam (Fig.~\ref{fig:beamwalking}). 
The resulting behavior is consistent with the other investigations: increasing duty factor improves the ability to traverse narrow terrain. 
We demonstrate that appropriate duty factor selection enables locomotion through constrained terrain across different velocities without adding specialized balance hardware such as a tail \cite{yang-proprioception-2023,huang_tail_2024} or reaction wheel \cite{lee_balance_2023}. 
These experiments provide a practical demonstration of how understanding gait parameters can be used to improve locomotion robustness through parameter selection rather than changing hardware or controller architecture.

The contributions of this work are:
\begin{enumerate}
\item A systematic analysis of the effects of gait type, duty factor, speed, and stance width on quadrupedal locomotion, showing that duty factor is a stronger predictor of local error convergence than gait type.
\item Evidence that this relationship persists across distinct planning and control approaches, robot models, and evaluation environments.
\item A novel demonstration of a quadrupedal robot traversing a narrow beam without the aid of additional balancing components such as a tail or reaction wheel.
\end{enumerate}

More broadly, these results suggest that gait type alone is an incomplete abstraction for understanding robust quadrupedal locomotion. Duty factor provides a simple parameter through which locomotion can instead be compared across gait types, evaluated under environmental constraints, and adapted to improve robustness.

\section{Related Works}
\label{sec:related}

\subsection{Gait Parameters in Biological and Robotic Locomotion}

A substantial body of work has investigated how gait parameters influence the performance of biological and robotic legged systems. In human locomotion, walking speed has been related to step length and step frequency through energetic considerations \cite{kuo_speed_step_2001,bertram_frequency_2001}, while step width has also been linked to energetic efficiency \cite{donelan_human_step_width_2001}. Other studies have examined how stride frequency, step width, and walking speed affect locomotion stability \cite{holt_stride_2010,young_step_stability_2012,england_speed_stability_2007}. Similar relationships have been observed in quadrupedal systems. Studies of horses have shown that different locomotion regimes are energetically advantageous at different speeds and that speed is correlated with stride frequency and duty factor \cite{hoyt_horse_1981,hoyt_energetics_2006}. 

In robotic quadrupeds, duty factor and other gait parameters have likewise been studied in relation to static stability and energetic efficiency \cite{mcghee_stability_1968,lee_gait_control_1986,koo_gait_stability_1999,mcclain_parameters_2018}. Collectively, this literature demonstrates that locomotion performance depends on a range of interacting gait parameters, including speed, frequency, step length, stance width, and duty factor.

Despite this broader parameter space, gait type remains a common abstraction for describing and comparing quadrupedal locomotion. 
%Walking, trotting, pacing, and bounding are typically distinguished by their inter-leg contact sequences, and gait type is often used to characterize differences in locomotion performance. 
In particular, prior studies of both biological and robotic quadrupeds have associated walking with efficient locomotion at lower speeds and trotting with improved efficiency or performance at higher speeds \cite{hoyt_horse_1981,Xi_gait_selection_2016,fu_emergent_gaits_2022}.
Methods have also been developed for robots to change gait types in order to improve locomotion stability \cite{humphreys2025learning,shafiee2024viability}.
This framing implicitly treats gait type as an important predictor of locomotion performance. However, gait type does not uniquely determine the temporal characteristics of locomotion: a given gait can be executed over a range of duty factors, speeds, stance widths, and gait periods. 
Consequently, comparisons between gait types can conflate the effects of contact sequence with those of the continuous parameters used to realize each gait.

\begin{table*}[t]
    \begin{center}
        \begin{tabular}{cccc} 
            Investigation & Planning/Control Architecture & Robot Model & Evaluation Environment\\
            \thickhline
            1 & Whole-Body Optimization w/ LQR Feedback & Ghost Spirit 40 & Pinocchio \\
            2 & Gait-Conditioned RL & Unitree Go2 & NVIDIA Isaac Sim \\
            3 & Centroidal MPC & Unitree Go2 & Hardware
        \end{tabular}
        \caption{Across three control architectures, two quadruped robot models, and three different evaluation environments in both simulation and hardware, we find consistent evidence that duty factor, as opposed to gait type, is a primary parameter that determines robustness of quadrupedal locomotion.}
        \label{table:investigation_overview}
    \end{center}
\end{table*}

\subsection{Locomotion on Narrow Terrain}

Narrow terrain presents a particularly challenging setting in which gait parameters can influence locomotion performance. As the available foothold region becomes narrower, the robot has less margin for maintaining its desired trajectory \cite{buchli_compliant_2009}. Prior work has demonstrated locomotion on narrow paths using specialized stabilization mechanisms, including control of the torso \cite{gonzalez_line_walking_2020}, and additional inertial elements such as tails \cite{yang-proprioception-2023,huang_tail_2024} or reaction wheels \cite{lee_balance_2023}. These approaches demonstrate that quadrupedal robots can successfully traverse highly constrained terrain, but often rely on additional mechanisms or specialized balance strategies. In contrast, we investigate whether appropriate selection of gait parameters can improve locomotion performance on narrow terrain without requiring these changes.

%We investigate whether continuous gait parameters provide a more informative description of quadrupedal locomotion than nominal gait type. In particular, we focus on duty factor and examine its relationship with convergence, energetic efficiency, and performance under constrained terrain. We evaluate these relationships across walking and trotting gaits in narrow-terrain environments, and across multiple planning and control approaches, providing evidence that duty factor can serve as a common parameter for comparing and adapting locomotion across nominal gait types (Table~\ref{table:investigation_overview}).

\begin{comment}
\begin{itemize}
    \item Early works on  quadruped walkers leveraged static stability and its relationship with contact sequence \cite{mcghee_stability_1968} and duty factor \cite{lee_gait_control_1986}.
    \item \cite{koo_gait_stability_1999} finds that higher duty factors corresponds to better stability for both static gaits (walk) and dynamic gaits (trot), though walk still a little better at similar duty factors. Reducing gait period also helps with stability.
\end{itemize}
\end{comment}

\begin{comment}
\section{Beam Walking}
\begin{itemize}
    \item A reaction wheel enabled a quarduped robot to navigate very narrow beams \cite{lee_balance_2023}. Recent work adding tails to quardrupedal robots could yield similar results \cite{yang-proprioception-2023}.
    \item \cite{gonzalez_line_walking_2020} demonstrates a quadruped balancing on two point feet, enabling narrow line walking.
\end{itemize}
\end{comment}

\section{Preliminaries}
\label{sec:prelims}

\subsection{Hybrid Systems}
\label{sec:hybrid_systems}

Legged locomotion is modeled as a hybrid dynamical system consisting of continuous dynamics within each contact mode and discrete transitions between modes. Let the continuous state be $x \in \mathcal{X}$ and the control input be $u \in \mathcal{U}$. Within a contact mode $I$, the system evolves according to
%\begin{equation}
$\dot{x} = f_I(x,u)$,
%\end{equation}
where $f_I$ denotes the continuous dynamics associated with the active contact configuration.

A transition from mode $I$ to mode $J$ occurs when the state reaches a guard surface
$
g_{I,J}(x) = 0.
$
At the transition, the state may undergo a discrete reset
$
x^+ = R_{I,J}(x^-),
$
where $x^-$ and $x^+$ denote the states immediately before and after the transition, respectively. In legged locomotion, these transitions correspond to foot touchdown and liftoff events.

For a nominal trajectory consisting of continuous motion and a sequence of hybrid transitions, perturbations to the trajectory evolve according to the continuous-time variational dynamics within each mode and are modified by the saltation matrix at each event \cite{kong2023saltation}. This provides a local linear approximation of how perturbations propagate through the complete hybrid trajectory.

\subsection{System Definition}
\label{sec:robot}

We consider a 12-degree-of-freedom quadrupedal robot with three actuated joints per leg. The configuration is represented by the position and orientation of the robot body together with the joint positions,
$
q =
[
p_B\  
q_B\ 
\theta_J
]
\in \mathbb{R}^{19},
$
where $p_B \in \mathbb{R}^3$ denotes the body position, $q_B \in \mathbb{S}^3$ denotes the body orientation represented by a unit quaternion, and $\theta_J \in \mathbb{R}^{12}$ denotes the joint angles. The generalized velocity is
$
v =
[
v_B \
\omega_B \
\dot{\theta}_J
]
\in \mathbb{R}^{18},
$
where $v_B$ and $\omega_B$ are the body linear and angular velocities. The full state is therefore $x=[q,v]\in\mathbb{R}^{37}$.
The robot dynamics are modeled separately for each contact mode $I$ as
$\dot{x} = f_I(x,u)$,
where $u\in\mathbb{R}^{12}$ denotes the control input.
In this work, we rely on existing simulators like Pinocchio \cite{carpentier2019pinocchio} and NVIDIA Isaac Sim \cite{gao_isaacsim_2026} to compute these hybrid dynamics.

For local convergence analysis, perturbations in orientation are represented in the three-dimensional tangent space of the unit quaternion rather than by direct subtraction of quaternion coordinates. Specifically, the orientation perturbation is represented by a local rotation vector $\delta\alpha_B\in\mathbb{R}^3$. The resulting local perturbation vector is
\begin{equation}
\delta x =
\begin{bmatrix}
\delta p_B &
\delta\alpha_B &
\delta q_J &
\delta v_B &
\delta\omega_B &
\delta\dot{q}_J
\end{bmatrix}^{T}
\in\mathbb{R}^{36}.
\end{equation}

Because these state components have different physical units and characteristic magnitudes, convergence is evaluated using a normalized state metric. Let $S$ be the diagonal scaling matrix corresponding to the state normalization used in the analysis. The scaled perturbation is then
%\begin{equation}
$\delta\bar{x}=S\delta x$.
%\end{equation}

\subsection{Convergence Analysis}
\label{sec:convergence}

We quantify local error convergence by linearizing the closed-loop hybrid dynamics about a nominal locomotion trajectory. Within a contact mode $I$, the continuous dynamics are linearized as,
%\begin{equation}
$\delta\dot{x}
=
A_I\delta x + B_I\delta u$.
%\end{equation}

For the trajectory-optimized controller, we use time-varying LQR feedback of the form, $\delta u = -K_I\delta x$,
giving the closed-loop variational dynamics,
%\begin{equation}
$\delta\dot{x}
=
(A_I-B_IK_I)\delta x$.
%\end{equation}
After discretization over timestep $h$, the perturbation evolves approximately according to
$\delta x_{i+1}
=
A^{\mathrm{cl}}_i\delta x_i,
$
where $A^{\mathrm{cl}}_i$ denotes the discrete-time closed-loop state transition matrix.

At a hybrid event, such as a foot touchdown or liftoff, the linearized perturbation is transformed by the corresponding saltation matrix, $\delta x^+ = \Xi\delta x^-$.
The saltation matrix accounts for the change in the continuous dynamics and the perturbation of the event timing caused by a state perturbation. 
It is dependent on the Jacobian of the guard and reset maps, and its derivation can be found in \cite{kong2023saltation}.
Combining the continuous-time and discrete event dynamics gives the linearized mapping from the initial to final perturbation along the complete nominal trajectory,
\begin{equation}
\Phi =
A^{\mathrm{cl}}_N
\Xi_{N-1}
A^{\mathrm{cl}}_{N-1}
\cdots
\Xi_1
A^{\mathrm{cl}}_1.
\end{equation}

Just as
%To account for the different units and characteristic magnitudes of the state variables, we evaluate this mapping using 
the normalized perturbation was $\delta\bar{x}=S\delta x$, with $S$ as a diagonal state-scaling matrix, the corresponding normalized state transition matrix is
%\begin{equation}
$\bar{\Phi}=S\Phi S^{-1}$.
%\end{equation}

Finally, the convergence metric is the largest singular value of this normalized transition matrix \cite{zhu2023convergent},
%\begin{equation}
$\chi = \sigma_{\max}(\bar{\Phi})$.
%\end{equation}
Thus, $\chi$ represents the maximum amplification of an infinitesimal perturbation over the nominal trajectory under the selected state metric. Values of $\chi<1$ indicate local contraction of all infinitesimal perturbations over the trajectory, while $\chi>1$ indicates that at least one perturbation direction is locally amplified.

\section{Methods}
\label{sec:methods}

We present three investigations that provide evidence that duty factor is a key predictor of quadrupedal locomotion robustness, particularly in challenging narrow environments.
%This finding challenges the conventional emphasis on gait type as a primary descriptor of locomotion performance. 
%Investigation 1 examines this relationship using whole-body trajectory optimization with LQR feedback, allowing us to systematically vary gait parameters while directly analyzing the resulting closed-loop error convergence.
%Investigation 2 applies these insights to a reinforcement learning controller, showing that adjusting duty factor also improves the performance of learned controllers in narrow-terrain environments.
%Finally, Investigation 3 demonstrates these relationships on a quadrupedal hardware platform.

\subsection{Investigation 1: Whole-Body Trajectory Optimization with LQR Feedback}
\label{sec:inv_1}

In Investigation 1, we use whole-body direct collocation trajectory optimization with LQR feedback to systematically study how gait parameters affect local error convergence and energetic efficiency. This approach provides direct control over the contact sequence and timing of each gait, allowing us to independently vary duty factor, forward speed, and stance width while maintaining a specified nominal gait type. 
%We consider both walking and trotting gaits to determine whether differences in performance are primarily associated with gait type or with the continuous parameters used to realize each gait.

We developed a custom direct collocation trajectory optimization framework using Pinocchio \cite{carpentier2019pinocchio} to compute the continuous and hybrid dynamics and their corresponding derivatives. %Direct collocation allows the contact sequence and timing to be explicitly specified, which is particularly useful for controlling gait type and duty factor. The resulting dynamics derivatives are also used to compute the fundamental solution matrix introduced in Section~\ref{sec:prelims}, which provides our measure of local error convergence.
The optimization objective consists of quadratic costs on control inputs together with shaping costs on body height and leg joint velocities.
%\begin{align}
%\quad & \sum_{i=0}^{N-1} u_i^\top R u_i
%+ x_{z,i}^\top Q_z x_{z,i}
%+ x_{lv,i}^\top Q_{lv} x_{lv,i},
%\label{eq}
%\end{align}
%where $X := {x_0,x_1,\ldots,x_N}$ is the sequence of system states, with $x_i \in \mathbb{R}^n$, and $U := {u_0,u_1,\ldots,u_{N-1}}$ is the sequence of control inputs, with $u_i \in \mathbb{R}^m$. The timestep length is denoted by $h$. The scalar $x_z$ denotes the state component corresponding to body height, while $x_{lv}$ denotes the vector of state components corresponding to leg joint velocities. $R$, $Q_z$, and $Q_{lv}$ are positive-definite weighting matrices of appropriate dimensions.
The trajectory is constrained to satisfy periodicity conditions on the internal state variables while translating forward at a specified velocity in the $x$ direction. 
%In particular, the final state is constrained to match the initial state up to the prescribed forward displacement over one gait cycle. 
The leg swing time is fixed, which combined with the specified duty factor, determines the gait period.
This gait period will differ across gait types and duty factors, but this approach yields better algorithm convergence than a fixed gait period.
%The peak swing height of each foot is also constrained to provide sufficient ground clearance. 
For this investigation, the robot model parameters were chosen to match the Ghost Robotics Spirit 40 quadruped.
%Soft constraints are applied to the initial position and velocity.
%This formulation therefore allows gait period, stance duration, and swing duration to vary with the selected duty factor and speed rather than imposing a fixed period across all trajectories. 
The complete formulation is omitted for brevity and will be provided with the open-source implementation accompanying this work.

For each optimized trajectory, we generate an LQR controller using fixed state and input weighting matrices. The resulting feedback controller stabilizes the nominal trajectory and enables us to evaluate local error convergence under closed-loop dynamics.
Given an optimal trajectory and LQR feedback controller, the convergence metric $\chi$ is computed as described in Sec.~\ref{sec:convergence}.

The state-scaling matrix $S$ is defined according to chosen characteristic error magnitudes corresponding to physically meaningful disturbances for the robot. 
We use characteristic scales of $0.05$~m, $0.1$~rad, $0.2$~rad, $0.1$~m/s, $0.1$~rad/s, and $1$~rad/s for body position, body orientation, joint position, body linear velocity, body angular velocity, and joint velocity, respectively, giving
\begin{equation}
S =
\operatorname{diag}
\left(
\frac{1}{0.05}I_3,
\frac{1}{0.1}I_3,
\frac{1}{0.2}I_{12},
\frac{1}{0.1}I_3,
\frac{1}{0.1}I_3,
I_{12}
\right)
\end{equation}
%where $I_N$ is an identity matrix of dimension $N\times N$.
While this choice of $S$ does affect the value of the convergence metric $\chi$, we find that the trends in Investigation 1 hold across reasonable choices of $S$.

\subsection{Investigation 2: Gait-Conditioned RL Locomotion Policy}
\label{sec:rl_policy}

In Investigation 2, we examine whether the relationships between gait parameters and locomotion performance extend to a learned locomotion controller for navigating narrow terrain.
We also explore whether duty factor can serve as a low-dimensional parameter for adapting learned locomotion.

Our framework consists of a gait-conditioned locomotion controller and a duty-factor selector. The locomotion controller executes commanded forward speed, duty factor, stance width, and gait period for the specified gait. The selector then learns to choose a duty factor based on measured success under external disturbances. 

We trained the gait-conditioned locomotion controller for the Unitree Go2 using Proximal Policy Optimization (PPO) in Isaac Lab \cite{mittal2025isaaclab}. The controller is a combination of separate trotting and walking networks with a common network architecture, observation and action interfaces, and reward formulation. Gait identity determines which policy is used, while the remaining gait parameters condition its behavior:  
\begin{equation}
    a_t = \pi_{\theta_g}(o_t,\phi_t,c_t),
    \qquad
    c_t = [v,d,w,T,g],
\end{equation}
where $a_t$ is the vector of joint-position actions, $o_t$ contains the robot observations, $\phi_t$ denotes the per-leg gait phases, and $\theta_g$ denotes the parameters of the policy for gait $g$. The command vector $c_t$ specifies forward speed $v$, duty factor $d$, full stance width $w$, gait period $T$, and gait type $g$.

The policy receives body-frame linear and angular velocities from the simulator, projected gravity, joint positions and velocities, previous actions, gait-phase encodings, gait commands, desired contact states, heading error, and measured foot contacts. The observation contains 68 inputs, and the policy produces 12 joint-position actions. 
The actor and critic each use a multilayer perceptron with hidden-layer widths of 256, 128, and 128 and ELU activations. Control operates at 50~Hz, while physics is simulated at 200~Hz. Joint-position targets are executed using nominal PD gains of $K_p=25$ and $K_d=0.5$. 

%The gait command specifies desired contact timing through leg-phase offsets and commanded duty factor. In front-left, front-right, rear-left, and rear-right order, the phase offsets are $(0,0.5,0.5,0)$ for trot and $(0,0.75,0.5,0.25)$ for walk. For leg $i$, the desired contact state is 
%\begin{equation}
%    b_i^{\mathrm{des}}(t)
%    =
%    \mathbb{I}\!\left[\phi_i(t)%<d\right],
%\end{equation}
%where $\phi_i(t)\in[0,1)$ is the leg phase. The phase encoding supplied to the policy is warped so that the commanded stance-to-swing transition occurs at a consistent location in the encoded cycle. Desired contacts and rewards remain defined using the original gait phase. 

The reward encourages forward-speed tracking, agreement with the desired contact schedule, appropriate lateral and fore-aft foot placement, swing-foot clearance, and body-height regulation. Additional penalties discourage heading errors and lateral drift. 
%Full stance width $w$ defines lateral foot-placement targets of $+w/2$ and $-w/2$ relative to the body. 
%The lateral placement reward uses a tolerance scale of 0.03~m to distinguish neighboring width commands. 
Self-collision is enabled to prevent leg interpenetration at narrow stances. 

Training covers randomized forward speeds of 0.25--0.40~m/s and stance widths of 0.05--0.30~m for both gaits. 
%Speed anchors are spaced at 0.05~m/s, and stance-width anchors are spaced at 0.05~m. 
Trotting duty factors span 0.50--0.75 and walking duty factors span 0.75--0.90. Gait periods range from 0.36--0.54~s for trot and 0.40--0.54~s for walk. 
%Duty-factor commands are constrained by minimum requested swing durations of five control steps for trot and two for walk:
%\begin{equation}
%    (1-d)T \geq %n_{\mathrm{sw},g}\Delta t,
%\end{equation}
%where $\Delta t=0.02$~s and $n_{\mathrm{sw},g}$ is the gait-specific minimum. Consequently, the feasible duty-factor range depends on period. For example, trotting at duty factor 0.75 is excluded for periods below 0.40~s.
Each gait network is trained for 1,800 PPO updates using 3,072 parallel environments and 48-step rollouts. A curriculum progressively expands the command distribution from a reference speed, width, and period to the full training domain. %Commands change only at gait-cycle boundaries. Training uses nominal phsyical parameter randomization. 
%Controller weights are fixed during evaluation. 

%The duty-factor selector operates above the locomotion controller. 
% Given stance width, forward speed, gait periood, and gait type, the duty-factor selector chooses a duty-factor command: 
% \begin{equation}
%     d = f_\psi(w,v,T,g).
% \end{equation}
% The selected command is supplied to the locomotion controller together with the remaining gait parameters. 
% %The selector does not generate joint actions or update the weights of the locomotion-policy. 

% We construct supervised targets by evaluating each feasible duty-factor candidate in 32 matched rollouts per context $x=[w,v,T,g]$. Candidates are spaced at 0.05 over
% 0.50--0.75 for trot and 0.75--0.90 for walk. Eligibility requires at least 29 rollouts to satisfy gait-tracking, contact-pattern, and failure checks and yield a valid mechanical cost-of-transport measurement. We select
% \begin{equation}
%     d^*(x)=
%     \underset{d\in\mathcal{D}_{\mathrm{eligible}}(x)}
%     {\operatorname{arg\,min}}\;
%     \operatorname{median}
%     \left[\mathrm{CoT}^{+}\mid x,d\right],
% \end{equation}
% where $\mathcal{D}_{\mathrm{eligible}}(x)$ is the eligible candidate set and $\mathrm{CoT}^{+}$ is positive mechanical cost of transport. The median is computed across qualifying rollouts. A classifier with two 32-unit hidden layers and ELU activations learns these targets. Its outputs are restricted to candidates feasible for the requested gait and period.
The selector network maps stance width, forward speed, gait period, and gait type to a duty-factor command,
%\begin{equation}
$    d=f_\psi(w,v,T,g)$.
%\end{equation}
To provide ground truth duty factor commands for the selector, 64 evaluation rollouts were performed at each gait type--duty cycle combination $d$, and context $x=[w,v,T,g]$. The duty cycle values are sampled at 0.05 intervals over 0.50--0.75 for trot and 0.75--0.90 for walk.

Given a context $x$, let $\hat{p}(x)$ denote the value of the highest success rate over the set of all evaluated gaits $\mathcal{D}(x)$.
We choose our ground truth duty factor label $d^*(x)$ as the minimum duty factor that achieves at least 90\% of $\hat{p}(x)$:
%\begin{equation}
%    \hat{p}_{\mathrm{best}}(x)
%    =\max_{d\in\mathcal{D}(x)}\hat{p}(d\mid %x),
%\end{equation}
%where $\mathcal{D}(x)$ is the feasible candidate set. For contexts with %$\hat{p}_{\mathrm{best}}(x)\geq0.05$, the %target is
%\begin{equation}
%    d^*(x)=
%    \min\left\{
%        d\in\mathcal{D}(x):
%        \hat{p}(d\mid x)
%        \geq 0.9\,\hat{p}(x)
%    \right\}.
%\end{equation}
%This rule favors lower duty factors when they retain nearly all of the best observed robustness. 
If all success rates for a given context are below 5\%, the context is excluded from supervised training.

A classifier with two 32-unit hidden layers and ELU activations learns these ground truth duty factor labels. 
Its outputs are masked to feasible candidates for the requested gait and period. The selector changes only the duty-factor command; the underlying locomotion control network remains fixed.

\subsection{Investigation 3: Centroidal MPC}
\label{sec:quad-sdk}

In the final investigation, we evaluate whether the relationship between duty factor and locomotion robustness persists under a separate planning and control architecture and on physical hardware. We use the open-source Quad-SDK package \cite{abs:norby-quad-sdk-2022}, which employs a centroidal model-predictive controller (MPC) to plan ground-reaction forces (GRFs) that are subsequently converted into joint commands. Unlike the approach in Investigation 1, this approach plans locomotion using a reduced-order centroidal dynamics model and tracks the resulting motion with an MPC controller.

The structure of Quad-SDK allows gait parameters and simulation environments to be modified without changing the underlying planning and control architecture. It also provides an existing local footstep planner and centroidal MPC implementation, enabling us to evaluate the effects of gait parameters using a controller that is independently developed from the whole-body optimization framework in Investigation 1.

Quad-SDK consists of three modules: a global body planner, a local planner, and a robot driver. For these experiments, we directly command body velocities through twist inputs and therefore bypass the global body planner. The local planner determines valid foothold locations using a predetermined traversability map \cite{kim2019highly} and provides these footholds to the centroidal MPC, which computes the ground-reaction forces required to track the desired center-of-mass trajectory. The robot driver then converts the planned GRFs into joint-level commands using inverse kinematics and interfaces with the robot's motor controllers.

We evaluate gait parameters using Quad-SDK on a Unitree Go2. 
By varying gait type, duty factor, forward velocity, and the width of the available foothold region, we assess whether the trends observed in Investigation 1 persist under a distinct locomotion architecture and in physical deployment. 
In particular, these experiments test whether higher duty factors enable more robust locomotion over narrow terrain.

\section{Experiments and Results}

\subsection{Investigation 1}

In this investigation, we perform three separate experiments to characterize the relationship between gait parameters and locomotion performance. %Experiment 1--1 provides a systematic sweep over gait type, duty factor, and forward speed to identify qualitative trends in convergence. Experiment 1--2 then isolates the effect of gait type by comparing walking and trotting at a matched duty factor. Finally, Experiment 1--3 evaluates whether duty factor is a stronger predictor of convergence when gait type, speed, and stance width are varied simultaneously.

\subsubsection{Experiment 1--1: Parameter Sweep}

We first generate walking and trotting gaits over a range of forward speeds and duty factors. For each gait type, we consider six forward speeds of $0.25$, $0.5$, $1$, $1.5$, $2$, and $2.5$ m/s. We evaluate six duty factors for each gait:
\begin{align*}
\text{Walk: } &[0.75,0.792,0.821,0.844,0.861,0.875]\\
\text{Trot: } &[0.5,0.583,0.643,0.688,0.722,0.75]
\end{align*}

\begin{table}[t]
    \begin{center}
        \begin{tabular}{cccc} 
            Predictor & Estimate $\beta$ & Standard Error & $p$-value\\
            \thickhline
            Gait (Walk) & 0.3867 & 0.221 & 8.82 $\cdot$ 10$^{-2}$ \\
            Speed & 0.646 & 0.111 & \textbf{1.21 $\cdot$ 10$^{\bf{-6}}$} \\
            Stance Width & -0.312 & 0.112 & \textbf{8.42 $\cdot$ 10$^{\bf{-3}}$}
        \end{tabular}
        \caption{Linear regression predicting convergence for walking and trotting gaits at matched DF=0.75.}
        \label{table:convergence_matched_stats}
    \end{center}
\end{table}

The minimum duty factors of $0.75$ for walking and $0.5$ for trotting correspond to the lowest values that avoid a full aerial phase.
For each trajectory, we compute the convergence metric as described in Section~\ref{sec:prelims}.
Four trajectories at the two highest speeds and two highest duty factors fail to converge to a feasible solution and are omitted. 
%These failures emphasize the gait parameters can not only be designed to optimize convergence, but also feasibility among other objectives.

\begin{figure}[t]
    \centering
    \includegraphics[width=0.98\linewidth]{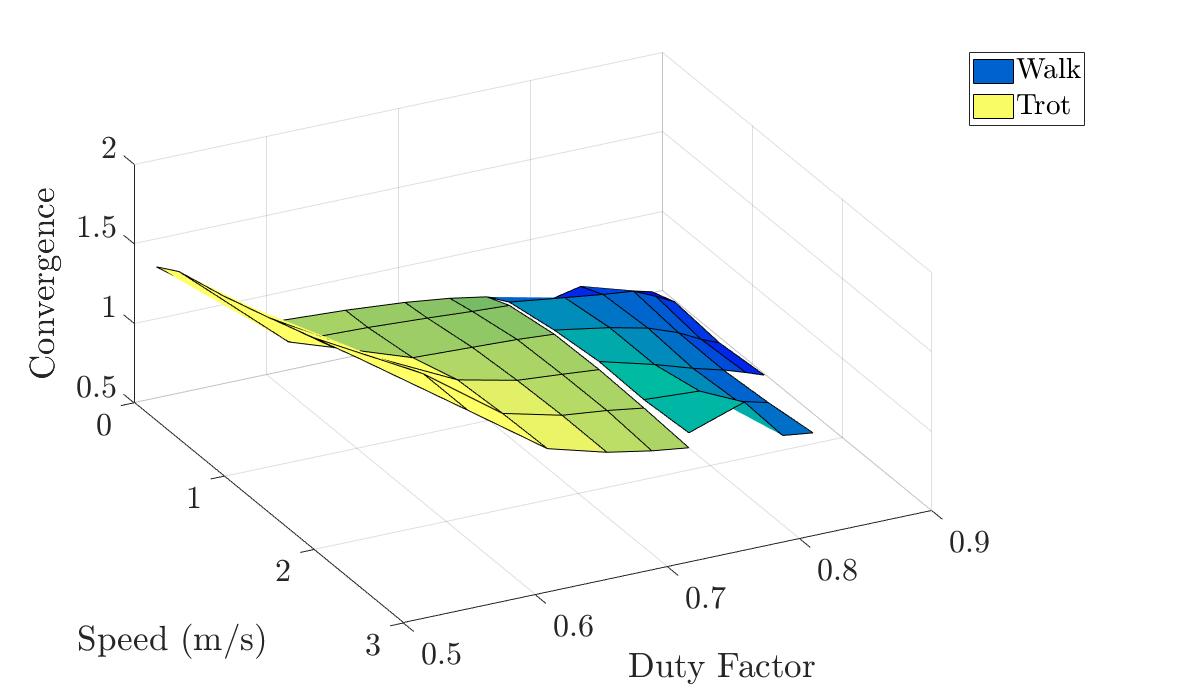}
    \caption{Convergence across walking and trotting gaits over a range of speeds and duty factors. Walking and trotting exhibit a continuous transition at the matched duty factor of $0.75$. Overall, convergence is strongly associated with duty factor.}
    \label{fig:plots}
\end{figure}

The results (Fig.~\ref{fig:plots}) reveal an approximately continuous transition between walking and trotting at the matched duty factor of $0.75$. 
%Although walking generally exhibits lower cost of transport at lower speeds and trotting becomes more efficient at higher speeds, these differences occur alongside systematic changes in duty factor. 
%Across both gait types, lower duty factors tend to be more efficient at higher speeds, while higher duty factors tend to be more efficient at lower speeds. This suggests that duty factor and speed jointly determine energetic efficiency rather than gait type alone.
Visually, it appears that convergence improves with increased duty factor, with higher duty factors generally producing lower values of the convergence metric and therefore greater local error contraction. Forward speed exhibits a weaker relationship with convergence than duty factor. 
These observations motivate the controlled comparisons in the following experiments.

\subsubsection{Experiment 1--2: Matched Duty Factor}

This experiment isolates the effect of gait type by holding duty factor fixed. We generate 20 walking and 20 trotting gaits, all with a duty factor of $0.75$. For each trajectory, forward speed and stance width are independently sampled from uniform distributions over $(0,3.0)$ m/s and $(0,0.35)$ m, respectively. We then fit linear regression models predicting convergence from gait type, forward speed, and stance width. 
All predictor variables are normalized before fitting the regression models.

\begin{comment}
\begin{table}[t]
    \begin{center}
        \begin{tabular}{cccc} 
            Predictor & Estimate $\beta$ & Standard Error & p-value\\
            \thickhline
            Gait (Walk) & 0.070 & 0.140 & 8.01 $\cdot$ 10$^{-1}$ \\
            Speed & 0.555 & 0.139 & 3.04 $\cdot$ 10$^{-4}$ \\
            Stance Width & -0.0260 & -0.186 & 8.54 $\cdot$ 10$^{-1}$
        \end{tabular}
        \caption{Linear regression predicting cost of transport for walking and trotting gaits at matched $DF=0.75$.}
        \label{table:CoT_matched_stats}
    \end{center}
\end{table}
\end{comment}

The regression results are summarized in Table~\ref{table:convergence_matched_stats}. 
For convergence, the walking gait has a positive coefficient relative to trotting, indicating higher predicted convergence values and therefore weaker local error contraction. 
However, this effect is not statistically significant at our threshold of $p<0.01$ ($p=0.0882$). In contrast, both forward speed ($p=1.21\times10^{-6}$) and stance width ($p=8.42\times10^{-3}$) are statistically significant predictors of convergence. 
Higher speeds and narrower stance widths are associated with higher convergence values and, therefore, worsened robustness.

%For cost of transport, gait type and stance width are not statistically significant predictors, while forward speed is significant ($p=3.04\times10^{-4}$). Thus, when duty factor is matched, we do not find strong evidence that gait type independently predicts either convergence or cost of transport.

This result is notable because walking and trotting use different contact sequences. Despite this, their convergence metrics are statistically similar when evaluated at the same duty factor, suggesting that contact sequence alone does not explain the observed differences in locomotion performance.

\subsubsection{Experiment 1--3: Duty Factor as a Predictor of Convergence}

Experiment 1--2 indicates that gait type does not explain convergence when duty factor is held fixed. 
Experiment 1--3 therefore investigates whether duty factor is a stronger predictor of convergence across gait types. 
We generate 30 additional trajectories for each gait type, with five trajectories generated at each of the 12 duty factors evaluated in Experiment 1--1, while speed and stance width are sampled from the same distributions used in Experiment 1--2. 
We fit a linear regression model using duty factor, forward speed, and stance width as predictors of convergence. 
%As before, all predictor variables are normalized before fitting the model.

\begin{table}[t]
    \begin{center}
        \begin{tabular}{cccc} 
            Predictor & Estimate $\beta$ & Standard Error & $p$-value\\
            \thickhline
            Duty Factor & -0.677 & 0.0863 & \textbf{1.42 $\cdot$ 10$^{\bf{-10}}$} \\
            Speed & 0.262 & 0.0853 & \textbf{3.29 $\cdot$ 10$^{\bf{-3}}$} \\
            Stance Width & -0.0621 & 0.0791 & 4.35 $\cdot$ 10$^{-1}$
        \end{tabular}
        \caption{Linear regression predicting convergence for walking and trotting gaits over a range of duty factors, speeds, and stance widths.}
        \label{table:Convergence_full_stats}
    \end{center}
\end{table}

The regression results are shown in Table~\ref{table:Convergence_full_stats}. 
Duty factor is a strong predictor of convergence, with a coefficient of $-0.677$ and a $p$-value of $1.42\times10^{-10}$. 
Because the predictor variables are normalized, the magnitude of this coefficient also provides a direct comparison with the effects of speed and stance width. The negative coefficient indicates that increasing duty factor decreases $\chi$ and results in stronger local error contraction.
The duty-factor coefficient has the largest magnitude among the three predictors, indicating that duty factor explains substantially more variation in convergence than either speed or stance width. 
%The negative coefficient indicates that increasing duty factor decreases the convergence metric, corresponding to greater local error contraction.

Forward speed is also statistically significant ($p=3.29\times10^{-3}$), with higher speeds associated with higher convergence values and therefore weaker local error contraction. 
Stance width is not statistically significant in this regression ($p=0.435$). Together, these results indicate that duty factor is the strongest linear predictor of convergence among the gait parameters considered in this experiment.

%We also fit a linear regression model to cost of transport. No individual predictor exhibits a strong linear relationship with cost of transport. However, the interaction between duty factor and forward speed is strongly associated with cost of transport, consistent with the trend observed in Fig.~\ref{fig:efficiency}. 
%This result further suggests that energetic efficiency depends on the combination of duty factor and speed rather than on gait type alone.

Overall, the three experiments provide complementary evidence that duty factor is a more informative parameter for characterizing locomotion robustness than nominal gait type. 
The initial parameter sweep indicates equivalent convergence across gait types at a matched duty factor. 
%At this matched DF, walking and trotting exhibit statistically similar convergence despite their different contact sequences. 
When duty factor is varied across both gait types, it emerges as the strongest predictor of convergence among the parameters considered. These results support treating duty factor as a continuous parameter for analyzing and selecting robust quadrupedal locomotion rather than relying solely on discrete gait type.

\subsection{Investigation 2}
\label{sec:rl_evaluation} 

%Experiments are conducted on flat ground to allow the stance width to be varied independently of terrain geometry. This enables controlled comparisons of gait execution, energetic efficiency, periodicity, and disturbance response across foot-placement configurations. Stance width denotes the commanded foot separation rather than physical support width.  

In this investigation, we conducted two experiments within the NVIDIA Isaac Sim simulator to evaluate: (1) the performance of the learned locomotion policy over commanded duty factors and stance widths; and (2) the relationship between stance width and the duty factor chosen by the selector network.

\subsubsection{Experiment 2--1: Duty Factor and Robustness}
We evaluated walking and trotting at widths of 0.05--0.30~m in 0.05~m increments. Duty factor commands ranged from 0.50--0.75 for trot and 0.75-0.90 for walk in 0.5 increments.% subject to the minimum swing-duration constraint. %Undisturbed evaluations confirm that realized duty factor increases with its command for both gaits. 

\begin{figure}[t]
    \centering
    \includegraphics[width=0.82\linewidth]
    {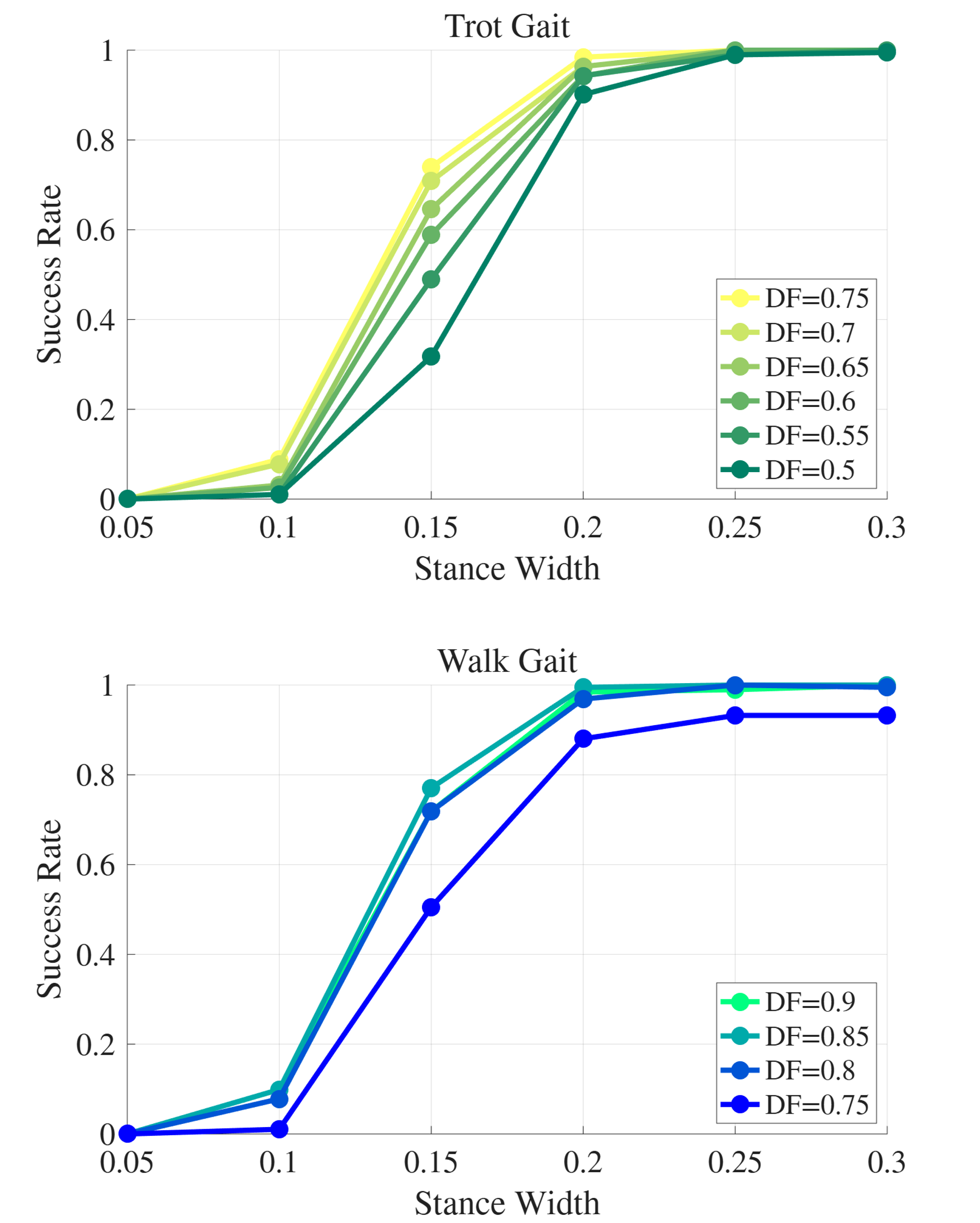}
    \caption{Across both trot and walk gaits, success rate of simulated trials increases with duty factor.}
    \label{fig:rl_robustness}
\end{figure}

To assess robustness as the robot traversed across the 3~m beam terrain, we applied randomized forces to the robot body of up to 25~N and torques of up to 3~N\,m. Disturbances were held for 0.10--0.40~s before resampling, with their magnitude envelope increasing over the first 10~s.  Figure~\ref{fig:rl_robustness} shows results at a commanded speed of 0.3~m/s. For each condition, 64 trials were conducted for gait periods of 0.40, 0.48, and 0.54~s, giving 192 trials per duty-factor/stance-width combination for each gait type.

Increasing commanded duty factor generally improved robustness at constrained stance width. The effect is strongest at the stance width of 0.15, where the trot gait with 0.5 DF was successful in just 31.8\% of trials. On the other hand, the success rate of the walk gait was over 70\% at duty factors of 0.8, 0.85, and 0.9.
%At wider stances, performance approaches a ceiling.
At the 0.05 stance width, none of the trials produced any successes.

%Higher duty factors improve traversal success at constrained stance widths. At 0.05~m, no evaluated duty factor succeeds. At 0.10~m, increasing duty factor raises trot success from 0\% at 0.50 to 7.8\% at 0.75, and walking success from 1.0\% at 0.75 to 9.9\% at 0.85.

%The largest improvements occur at intermediate widths. At 0.15~m, trot success increases from 23.4\% at duty factor 0.50 to 74.2\% at 0.75, while walking success increases from 50.5\% at 0.75 to 77.1\% at 0.85. At 0.20~m, the corresponding increases are 87.5\% to 97.7\% for trot and 88.0\% to 99.5\% for walk. At 0.25-0.30~m success remains sustainably high, ranging from 98.4--100\% for trot and 93.2--100\% for walk. 

\subsubsection{Experiment 2--2: Duty Factor Selection Network}
We constructed selector targets across four forward speeds (0.25, 0.30, 0.35, and 0.40~m/s), five stance widths (0.05--0.3~m), and gait periods of 0.36, 0.40, 0.48, and 0.54~s for trot and 0.40, 0.48, and 0.54~s for walk. Each combination defines one operating condition, giving $168$ conditions. Within each condition, feasible duty-factor candidates are compared to construct the selector target. Of the 168 contexts, 133 satisfy the labeling criterion: 76 for trot and 57 for walk. The remaining 35 contexts had success rates below 5\% and were excluded from training. %The classifier reproduces all 133 targets exactly.

\begin{figure}[t]
    \centering
    \includegraphics[width=0.85\linewidth]
    {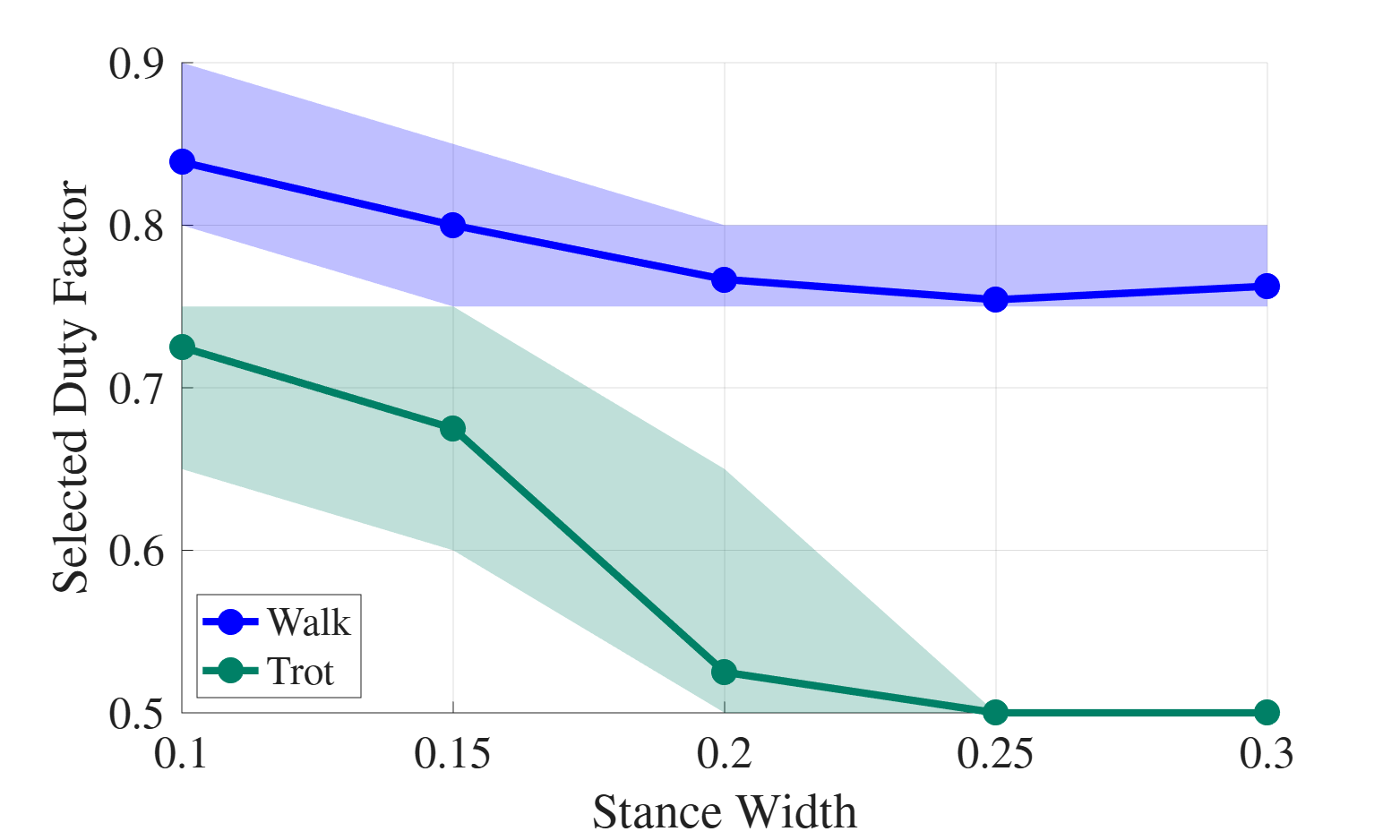}
    \caption{The duty factor selector network shows a trend of selecting higher duty factors as stance width narrows. Plot shows mean values and shaded regions represent minimum and maximum chosen values at each stance width.}
    \label{fig:rl_selector}
\end{figure}

Figure~\ref{fig:rl_selector} summarizes the network's duty factor selections as stance width changes. For the trot, mean selected duty factor decreases from 0.725 at a stance width of 0.10~m to 0.500 at the widest stance widths. For the walk, the mean duty factor is 0.839 at the 0.10~m width and just over the minimum of 0.75 at wider widths. 
No context is supported at 0.05~m, so this stance width is omitted from the figure. 
Note that the control policy still exhibits low success rates at 0.10~m, demonstrating that the relative selection criterion does not ensure reliable locomotion.
%At 0.10~m, the figure includes only the eight supported trotting contexts and nine supported walking contexts out of 12 per gait.

%We evaluate each supported selection using 64 fresh rollouts under the same disturbance distribution, giving 8,512 validation trials. Pooled across supported speed and period contexts, success at widths of 0.10, 0.15, and 0.20~m is respectively 27.0\%, 75.5\%, and 92.8\% for trot, and 26.0\%, 80.1\% and 96.0\% for walk. At 0.25--0.30~m, success reaches 99.8--100\% for trot and 96.5--97.5\% for walk. 

These results show that the selector generally chooses higher duty factors at narrower stances while selecting lower values where comparable robustness is available. 
This aligns with the previous findings that duty factor and robustness are highly correlated, particularly as stance width is narrowed.
%Even under different robot models, control architectures, simulation environments, and evaluation methods, this pattern holds across the first two investigations.

\subsection{Investigation 3}

To evaluate the effect of speed, duty factor, and stance width on the locomotion performance of a physical robot, we built a custom narrow beam terrain with width 0.15 m (Fig.~\ref{fig:beamwalking}).
A Unitree Go2 quadruped robot was tasked with traversing these beam terrains with planning and control as described in Sec.~\ref{sec:quad-sdk}.

Nine sets of trials were run at three different gaits (trot at 0.5 DF, trot at 0.75 DF, and walk at 0.75 DF) and three commanded speeds 0.3, 0.4, and 0.5 m/s.
Each of these gait and speed combinations were evaluated on the 0.15 m width beam.
Each set of trials was executed 5 times, giving 45 total trials.
Each trial was evaluated on if the robot successfully traversed the entirety of the beam length without falling.

Table~\ref{table:hardware} shows the outcome of the hardware experiments.
For both of the higher duty factor gaits, the robot successfully traversed the beam 100\% of the trials.
On the other hand, the trot gait with 0.5 duty factor failed at each speed, completing a total of 12/15 trials.
In each of these three failures, the robot lost balance in the middle of the beam and fell off.

While this set of hardware experiments suggests a difference in performance between duty factors, the results do not show a vast gap in performance. 
Since the robot is highly dependent on accurate state estimation and sensing, the robot leveraged motion capture to estimate its position relative to the world.
Prior to execution of the trials, the motion capture system was carefully calibrated to ensure accuracy.
This precise calibration and estimation ability may have amplified the robot's ability to successfully traverse the beam, even at lower duty factors.
In future work, we would like to explore how noisier vision-based estimation methods may be more sensitive to duty factor in similar experiments.

\renewcommand{\arraystretch}{1.2}
\begin{table}[t]
    \begin{center}
        \begin{tabular}{ccccc} 
            Gait Type & Duty Factor & Speed (m/s) & Success Rate\\
            \thickhline
            \multirow{3}{*}{Trot} &
            \multirow{3}{*}{0.5}  & 0.3 & 4/5\\ 
             & & 0.4 &  4/5\\ 
             & & 0.5 &  4/5\\ 
            \hline
            
            \multirow{3}{*}{Trot} & \multirow{3}{*}{0.75} & 0.3 & \textbf{5/5}\\  
             &  & 0.4 & \textbf{5/5}\\ 
             & & 0.5 &  \textbf{5/5}\\  
            \hline
            \multirow{3}{*}{Walk} & \multirow{3}{*}{0.75} & 0.3 & \textbf{5/5}\\ 
            &  & 0.4 & \textbf{5/5}\\ 
            &  & 0.5 & \textbf{5/5}\\  
        \end{tabular}
        \caption{At the duty factor of 0.75, the trot and walk both achieved 100\% success on the hardware experiment. The lower duty factor trot experienced several failures.}
        \label{table:hardware}
    \end{center}
\end{table}

\section{Discussion \& Limitations}

The three investigations presented in this work provide converging evidence that duty factor is an important predictor of quadrupedal locomotion robustness, particularly when the robot must operate with constrained stance width. Nevertheless, the scope of these conclusions should be interpreted in light of several limitations.

First, isolating the effects of individual gait parameters requires holding other aspects of locomotion approximately constant, such as gait period, swing-leg timing and trajectory, and body height. 
These parameters may also affect locomotion performance and may interact with duty factor.
We therefore do not claim that duty factor is the sole determinant of locomotion robustness. 
Rather, it consistently provided substantially more explanatory power than gait type. 
A more complete characterization of these interactions is an important direction for future work.

Second, our analysis focuses on walking and trotting, which are among the most commonly used gait families in contemporary quadrupedal locomotion. 
We do not evaluate other gaits, such as bounding, pronking, or cantering. 
The relationship identified here should therefore not be interpreted as a universal statement that gait type is irrelevant to all forms of quadrupedal locomotion. 
Instead, our results demonstrate that even two gait types traditionally treated as fundamentally distinct do not necessarily yield distinct performance.

%Third, the narrow-terrain experiments address one particularly relevant challenge for real-world deployment, but quadrupedal robots must also tolerate disturbances arising from actuator failures, contact uncertainty, localization error, and more. 
%Future work should examine whether duty factor provides similarly useful predictive or adaptive value across these additional sources of uncertainty.

Despite these limitations, the consistency of the results across the three investigations provides strong evidence that gait type is not a good predictor of locomotion performance and duty factor is much more highly correlated.
The practical implication is that gait selection need not begin with a discrete choice between ``walk'' and ``trot''. 
Instead, duty factor can serve as a continuous and interpretable control parameter for adapting locomotion to task and environmental constraints. 
This perspective can simplify gait selection and provide a direct mechanism for increasing robustness when environmental constraints demand it.

\section{Conclusion}

This work investigated the role of duty factor in quadrupedal locomotion across different gait types, control architectures, and operating conditions. Across three investigations, we found consistent evidence that duty factor provides a more informative description of robustness than nominal gait type for the walking and trotting gaits studied. 
%Whole-body trajectory optimization with LQR feedback showed that walking and trotting exhibit similar convergence metrics when evaluated at matched duty factors, while duty factor was strongly associated with the resulting convergence behavior across a broader range of gaits. A learned locomotion controller further demonstrated that duty factor can be exposed as an explicit parameter for adapting locomotion behavior. Finally, experiments using an independent centroidal MPC architecture showed that higher duty factors improve traversal of increasingly constrained foothold regions in both simulation and hardware.

These results suggest that gait selection should not necessarily be treated as a discrete choice between predefined gait types. Instead, continuous gait parameters such as duty factor may provide a basis for selecting locomotion behavior according to environmental constraints and desired performance. Future work will investigate whether this relationship extends to additional gait parameters, locomotion regimes, and sources of uncertainty, as well as whether duty factor can be automatically adapted online to balance competing objectives like robustness and energetic efficiency during real-world locomotion.

\section{Acknowledgments}

ChatGPT was used for text editing while preparing this manuscript.
                                  
\bibliographystyle{IEEEtran}
\bibliography{ref}

% that's all folks
\end{document}